\documentclass[conference]{IEEEtran}
\ifCLASSINFOpdf
  \usepackage[pdftex]{graphicx}
\else
  \usepackage[dvips]{graphicx}
\fi
\usepackage{amsmath}
\ifCLASSOPTIONcompsoc
  \usepackage[caption=false,font=normalsize,labelfont=sf,textfont=sf]{subfig}
\else
  \usepackage[caption=false,font=footnotesize]{subfig}
\fi

\usepackage{stfloats}
\usepackage{url}
\usepackage{cite}
\usepackage{color}
\usepackage{gensymb}
\usepackage{comment}
\usepackage{multirow}
\usepackage{caption}

\usepackage[noend]{algpseudocode}
\usepackage{algorithm}
\usepackage[ruled,vlined,algo2e]{algorithm2e}
\providecommand{\SetAlgoLined}{\SetLine}

\begin{document}
%
\title{A Near Sensor Edge Computing System for \\ Point Cloud Semantic Segmentation}

\author{
\IEEEauthorblockN{Lin Bai, Yiming Zhao and Xinming Huang}
\IEEEauthorblockA{
Department of Electrical and Computer Engineering\\
Worcester Polytechnic Institute, Worcester, MA 01609, USA\\
\{lbai2, yzhao7, xhuang\}@wpi.edu}

}


\maketitle

\begin{abstract}
Point cloud semantic segmentation has attracted attentions due to its robustness to light condition. This makes it an ideal semantic solution for autonomous driving. However, considering the large computation burden and bandwidth demanding of neural networks, putting all the computing into vehicle Electronic Control Unit (ECU) is not efficient or practical. In this paper, we proposed a light weighted point cloud semantic segmentation network based on range view. Due to its simple pre-processing and standard convolution, it is efficient when running on deep learning accelerator like DPU. Furthermore, a near sensor computing system is built for autonomous vehicles. In this system, a FPGA-based deep learning accelerator core (DPU) is placed next to the LiDAR sensor, to perform point cloud pre-processing and segmentation neural network. By leaving only the post-processing step to ECU, this solution heavily alleviate the computation burden of ECU and consequently shortens the decision making and vehicles reaction latency. Our semantic segmentation network achieved 10 frame per second (fps) on Xilinx DPU with computation efficiency 42.5 GOP/W.

\end{abstract}

\section{INTRODUCTION}

Surrounding environment understanding is one of the essential tasks for autonomous vehicles. Semantic segmentation, by classifying each pixel of an image into corresponding class (such as cars, pedestrians, buildings or etc.) of what is being represented, gives autonomous vehicles an accurate abstraction for scene understanding.

Semantic segmentation on images has been well studied \cite{ronneberger2015u, long2015fully}. However, for autonomous vehicles, which has to run not only in day time but also at night, camera solution is not reliable enough. In recent year, LiDAR point cloud has been widely used for semantic segmentation task \cite{qi2017pointnet, qi2017pointnet++, landrieu2018large, hu2020randla, li2018pointcnn, wu2018squeezeseg, milioto2019rangenet++, wu2019squeezesegv2, wang2018pointseg, xu2020squeezesegv3, alonso20203d, lawin2017deep, liong2020amvnet, gerdzhev2020tornado, su2018splatnet, rosu2019latticenet}. By emitting photons in a rotatory way and locating the hit point using flying time, a LiDAR is able to work in low light and even no light scenarios. Different from the pixels in images, point cloud elements are represented in 3D Cartesian coordinate. Thus, a well-segmented point cloud gives richer location information than that from a well segmented image.

\begin{figure}[htbp]
\includegraphics[width=1.0\linewidth,height=2.5cm]{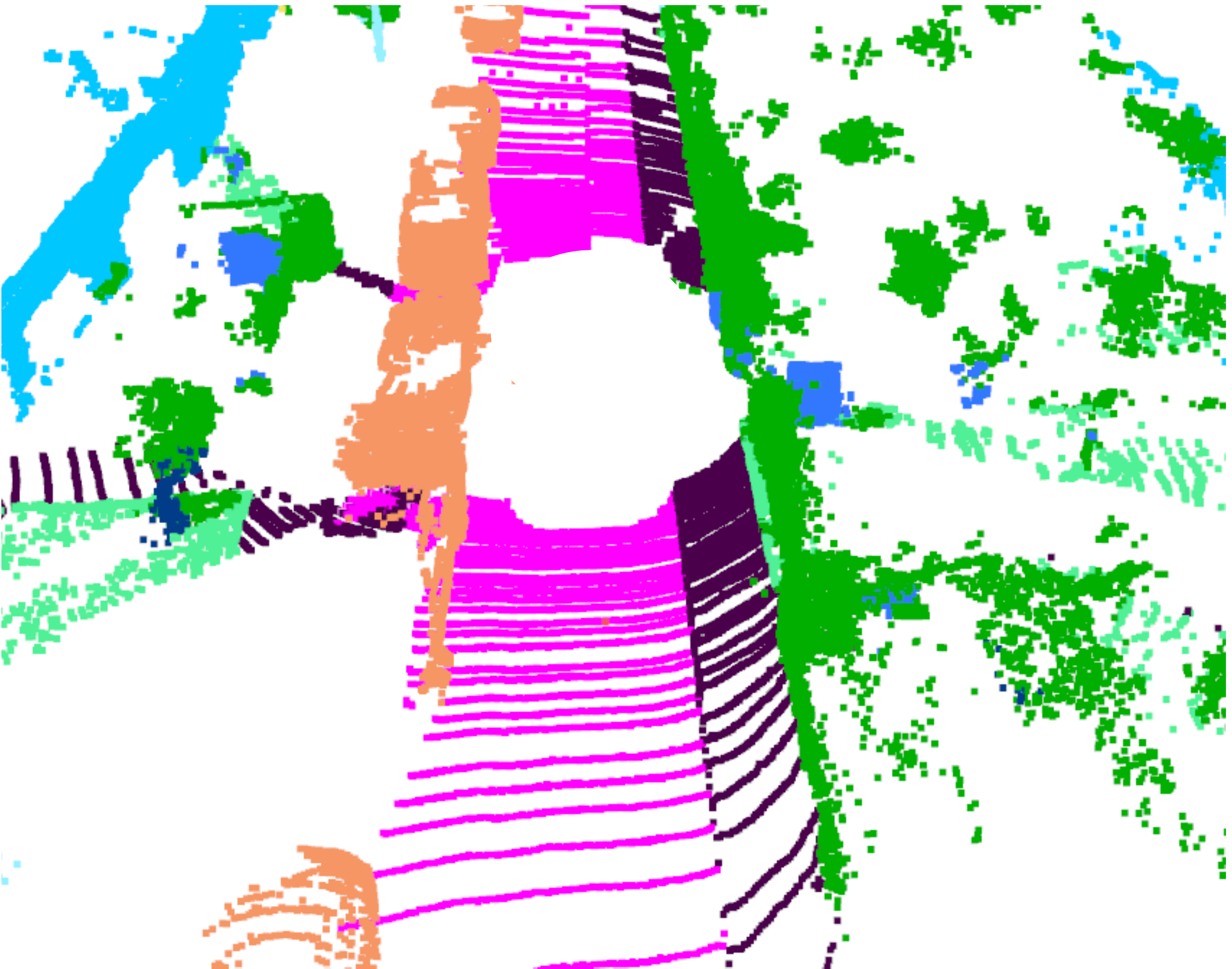}
     \caption{The LiDAR point cloud semantic segmentation predicts a semantic label for each point to help the car to understand the 3D surroundings. This is a sample from the validation sequence in the SemanticKITTI dataset.}
    \label{fig:fisrt}
\end{figure}

Another practical issue for autonomous driving application is the processing speed. A typical decision making chain is as follows. Data is firstly captured by LiDAR, and then sent into the semantic segmentation neural network for point-wise classification. This dense prediction is then transmitted into central processing unit for path planning and/or trajectory planning. After decision making, the mechanical part starts to execute this decision. Considering autonomous driving, such a time-critical task, it is necessary to squeeze the processing time of the semantic segmentation task in our case. A typical scanning rate for LiDAR is 100ms.

What's more, almost all the state-of-the-art (SOTA) point cloud semantic segmentation networks are targeting to GPUs, which are not suitable for edge computing. Due to the consideration on computation efficiency, edge deep learning accelerators (like Xilinx DPU \cite{xilinxDPU} and NVIDIA NVDLA \cite{nvidia}) do not support all the commonly used operations. Therefore, edge deep learning accelerator compatible networks are critical for real-time embedded applications.

In this paper, we propose a near sensor computing solution for light-weighted semantic segmentation. It addresses the aforementioned issues in two aspects:

(1) A light weighted semantic segmentation network targeting to DPU is proposed. By adopting bi-linear interpolation instead of deconvolution, and using hardware friendly layers, the computation burden is drastically reduced;

(2) A near sensor computing solution based on DPU is proposed. Through moving the semantic segmentation neural network from central processing unit into computation \& bandwidth efficient DPU, this near sensor solution alleviates the computation burden, and consequently results in faster decision making procedural.

\section{Related Work}
The existing point cloud semantic segmentation networks can be categorized into two groups: point-based networks and projection-based networks. These two methods require not only different network design but also different representation of the point cloud. Point-based methods compute raw point cloud directly, while projection-based approaches adopt various projection ways to map the 3D point cloud into 2D plane.

\subsection{Point-based networks}
PointNet \cite{qi2017pointnet} utilizes the MLP (multi-layer perceptron) based network to deal with point cloud. In this case, there is no requirement to point cloud order. To address the rotation variance issue, a learning based transformation module (T-net) is introduced. To reduce the heavy computation burden of PointNet, the subsequent network, PointNet++ \cite{qi2017pointnet++} processes the point cloud with K-Nearest Neighbour (KNN) layer. Besides, PointNet++ proposes novel set learning layers to adaptively combine feature from multi-scales. But the local query and grouping still limit the model performance on a large point cloud. Recent research work \cite{landrieu2018large, hu2020randla} solve these issues by adopting graph network. PointConvs, different from PointNet-like networks, introduces a special convolution kernel and shows a strong generality. Typical works like PointCNN \cite{li2018pointcnn} and KPConv \cite{thomas2019kpconv} are investigated for the semantic segmentation task. One problem of all the previous point-based networks is that, processing capability and memory requirement increase sharply as the point cloud becomes larger.

\subsection{Projection-based networks}
\textbf{Project to voxel:} A common way to project 3D point cloud into 2D is voxelization. It discretizes the 3D space into 3D volumetric space and assigns each point to the corresponding voxel \cite{zhang2018efficient, zhou2018voxelnet, tchapmi2017segcloud}. However, the sparsity and irregularity of the point cloud lead to redundant computations in voxelized data since many voxel cells may be empty, especially the points far away from the LiDAR. 

\textbf{Project to range view:} To conquer the redundant computation issue in voxel projection networks, spherical range view projection networks are proposed \cite{wu2018squeezeseg, milioto2019rangenet++, wu2019squeezesegv2, wang2018pointseg, xu2020squeezesegv3, alonso20203d}. Unlike point-wise and other projection-based methods, the 2D rendered image representations of range view based approach are more compact, dense and computationally cheaper, since standard 2D convolutional layers can process them.

\textbf{Others:} There are some other projection-based methods are investigated on point cloud semantic segmentation task, such as multi-view representation \cite{lawin2017deep, liong2020amvnet, gerdzhev2020tornado} and lattice structure \cite{su2018splatnet, rosu2019latticenet}.

Considering the simple project mechanism and computation efficiency on DPU, the range view projection approach is used in our network.

\section{Network Design}
As mentioned in the previous section, two mainstream approaches for point cloud semantic segmentation include point-based method and project-based method. Point-based method process the raw 3D point cloud directly. No transformation or pre-processing is involved. The projection-based method, however, transforms the 3D point cloud into various formats, such as voxel cells, multi-view representation, lattice structure, or rasterized images. In this paper, we choose to transform point cloud into range view in spherical coordinate. There are mainly three reasons: 1) Projection based method takes the advantage of convolutional neural network. Also it avoids many extra operations (like voxelization in RPVNet \cite{xu2021rpvnet}, cylinder partition in Cylinder3D \cite{zhu2021cylindrical} and K-Means in PointNet++ \cite{qi2017pointnet++}) or computation consuming operations (like MLP in PointNet); 2) The transformation process is simple. According to the mechanism of rotary LiDAR, the points are intrinsic in spherical coordinate; 3) The range view is dense. When projected into image view, the feature map would be very sparse. While the range view has valid values for almost every pixel on it. Thus, 2D convolutional neural networks can be applied for feature extraction in range view as well.

\begin{figure*}[htbp]
    \centering
    \includegraphics[width=0.85\textwidth]{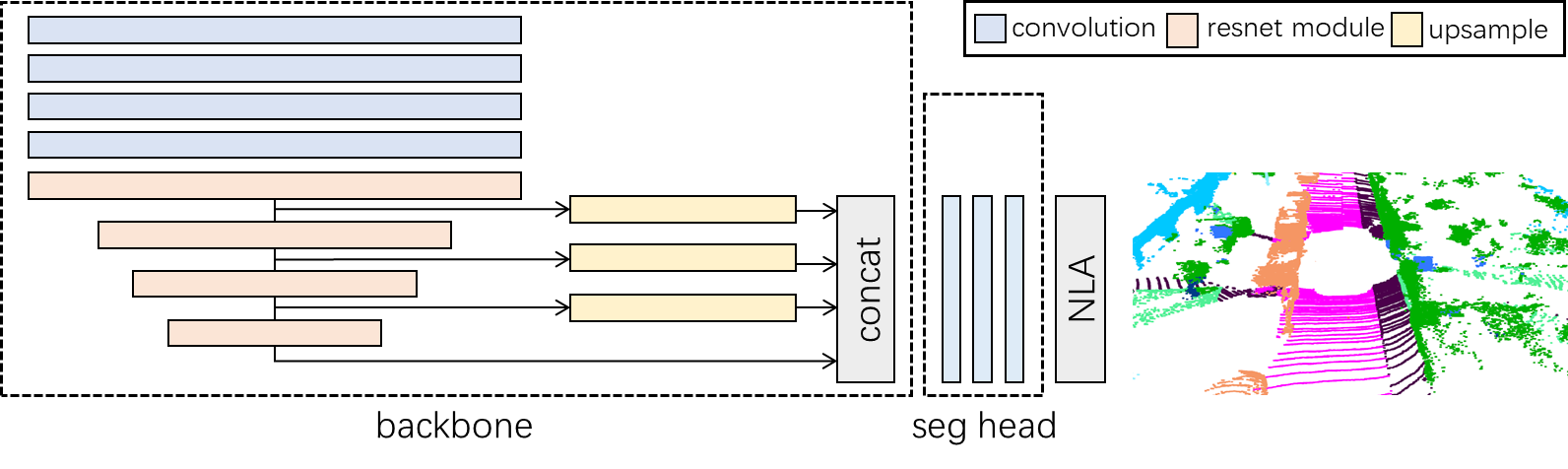}
    \vspace{-3mm}
    \caption{The diagram of our network}
    \label{fig:backboe}
\end{figure*}

\subsection{Input Module}
In spherical coordinate, each point is described in $(r, \theta, \phi)$, where $r$ is radial distance, $\theta$ is polar angle, and $\phi)$ is azimuthal angle. The range view projection maps each point from Caresian coordinate into spherical coordinate, a 2D $(\theta, \phi)$ map. Most of the projection-based neural networks use five channels $(x, y, z, r, remission)$ as the input. But as discussed in \cite{badino2011fast} and \cite{zhao2021surface}, adding a normal vector for each point can stablize the training process. Therefore, besides the $(x, y, z, r, remission)$ channels, three extra channels ($n_1$, $n_2$, and $n_3$) are concatenated into input feature map.

\subsection{Backbone}
The backbone module utilized in our network is ResNet-34. It has a better feature extraction capability than ResNet-18 but less computation than ResNet-50. Different from the standard ResNet-34 configuration, in order to extend the field of perception, we extend the convolution stride to 2 for the last three residual modules of ResNet-34. ASPP \cite{chen2017rethinking} is one of the most popular modules for semantic segmentation task. By applying convolutions with different dilated rates to the feature map, ASPP concatenates information with multiple fields of perception and consequently achieves good performance. However, from the hardware (GPU or DPU) point of view, the dilated convolution is not as efficient as standard convolution (dilated rate is 1) and results in slow inference speed. The convolution with larger dilated rate makes the processing time even worse. Therefore, we keep the dilated rate equaling to 2 for last three modules of ResNet-34 (Fig.~\ref{fig:backboe}), and then concatenate all of them. Considering the unequal sizes of the residual modules, the bi-linear interpolation is utilized for the classification head. By employing this approach, (1) the uniformed dilated rate to cascaded convolutions simulates the functionality of ASPP without decreasing the computation efficiency on GPU or DPU; (2) resize part is totally parameter-free, which speeds up the inference time.

\subsection{Classification Head}
The classification head in our network contains three convolutional layers with 1$\times$1 kernel. The mainstream high performance segmentation networks like U-net\cite{ronneberger2015u}, FCN\cite{long2015fully} like layers of devolution with skip connections. But this requires more computation and data movement on GPU or DPU. While the light weighted segmentation networks like BiSeNet\cite{yu2018bisenet} and ICNet\cite{zhao2018icnet} take the advantage of bi-linear interpolation to speed up the decoding part. In our case, a compromised way is adopted. After the decoding by the bi-linear interpolation layers in backbone, three extra layers of convolution are added at the end of it to refine the classification results. To further reduce the number of parameters and computation, all the convolution kernel size are unified to 1 $\times$ 1.

\subsection{Post-processing}
The output of segmentation networks may encounter boundary blurring effect. This is due to the many-to-one mapping on the range view image. Ideally, mapping from $(x, y, z)$ Cartesian coordinate to $(r, \theta, \phi)$ spherical coordinate is one-to-one continuous. However, the discretizing process from $(r, \theta, \phi)$ to 2D $(\theta, \phi)$ image may group more than one close points into one cell. However, only the information of one point in the cell will be processed by the neural network. Considering the points mapped in the same cell in the range view, they are close to each other on the range image, but actually they may be far away from each other in the the real world and belong to different categories (labels). The most common way to alleviate this effect is K-Nearest Neighbor (KNN) function. However, according to the experiment in \cite{zhao2021fidnet}, a simpler solution named Nearest Label Assignment (NLA) is enough for our case, which removes the Gaussian weighting step and range cutoff step. The algorithm details are shown in Alg.~\ref{alg:NLA}.

\begin{algorithm}[htbp]
	\SetAlgoLined
	\SetKwInOut{Input}{Input}\SetKwInOut{Output}{Output}
    \Input{Range image $I_{r}$ with size $H \times W$,\\
           predicted label map $I_{label}$ with size $H \times W$,\\
           vector $R_{all}$ with range values for all points,\\
           vector $h_{all}$ with projected $h$ values for all points,\\
           vector $w_{all}$ with projected $w$ values for all points,\\
           local kernel size $k$.
           }
	\Output{Vector $Labels$ with predicted labels for all points.}
	\BlankLine
	$Labels \leftarrow \hspace{2mm}empty \hspace{2mm} list \hspace{2mm} [\hspace{2mm}], \hspace{3mm} k \leftarrow 5 \\
	S(h,w,k) \leftarrow \forall (h_{n},w_{m}), \hspace{1mm}where\hspace{1mm} (h_{n},w_{m}) \hspace{1mm}in \hspace{1mm}the \hspace{1mm} k \times k \hspace{1mm} local\hspace{1mm} patch\hspace{1mm} centered \hspace{1mm} at\hspace{1mm} (h,w);$\\ 
	    \ForEach{$i$ in $1 : R_{all}.length()$}{
	    $min\_diff \leftarrow +\infty$;
	    \\
	    \ForEach{position $(h_{n},w_{m})$ in $S(h_{all}[i],w_{all}[i],k)$}{
	        \If{$abs(I_{r}(h_{n},w_{m})-R_{all}[i])<min\_diff$}{$label\_each=I_{label}(h_{n},w_{m})\\
	        min\_diff=abs(I_{r}(h_{n},w_{m})-R_{all}[i])$} 
	        
	        $Labels.append(label\_each)$
	    }
	    }
	\Return{$Labels$}
	\caption{Nearest Label Assignment (NLA) \cite{zhao2021fidnet}}\label{alg:NLA}
\end{algorithm}

\section{Network Training}
\subsection{Dataset}
Our network is trained on the SemanticKITTI dataset \cite{behley2019semantickitti}, which is a large-scale dataset that provides dense point-wise annotations for the entire KITTI Odometry Benchmark \cite{geiger2012we}. It consists of 22 sequences totally. Following the official split way, we train the model on sequences 00 to 07, plus 09 and 10. Sequence 08 is used as validation set. And the test set contains sequences 11 to 21.

\subsection{Training Setting}
The training data have been augmented following the methods used in other works \cite{gerdzhev2020tornado, cortinhal2020salsanext}, rotation along the $\gamma$ axis and flipping along the $\gamma$ axis.The loss function is a combination of the weighted cross-entropy loss from \cite{zhang2018generalized} and the Lovász-Softmax loss from \cite{berman2018lovasz}. The optimizer is Adam. And the learning rate decay follows a cosine annealing-like way. If using the mix-precision choice in PyTorch, the network fits NVIDIA QUADRO RTX 8000 with the batch size equaling to 24.

\subsection{Performance}
\begin{figure*}[htbp]
  \captionsetup[subfigure]{labelformat=empty}
  \subfloat[prediction of scan 0]{
	\begin{minipage}[c]{0.32\textwidth}
	   \centering
	   \includegraphics[width=1\textwidth,height=30mm]{{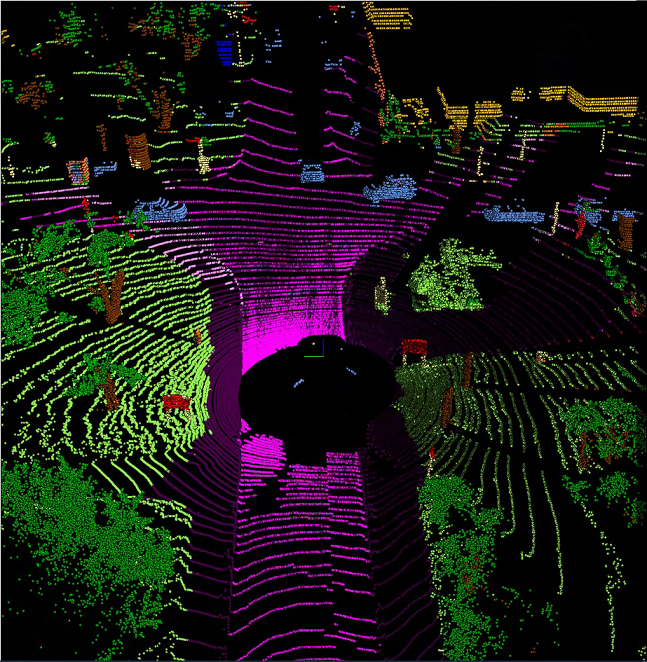}}
	\end{minipage}}
  \hfill
  \subfloat[prediction of scan 40]{
	\begin{minipage}[c]{0.32\textwidth}
	   \centering
	   \includegraphics[width=1\textwidth,height=30mm]{{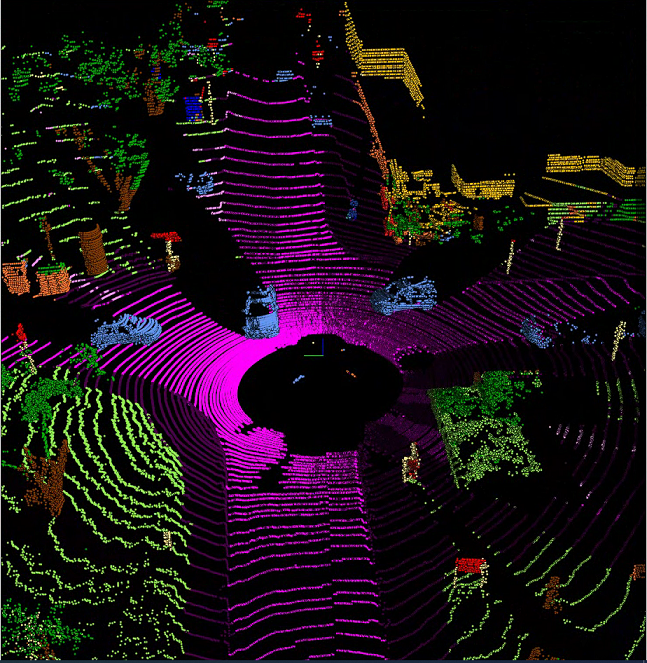}}
	\end{minipage}}
  \hfill
  \subfloat[prediction of scan 140]{
	\begin{minipage}[c]{0.32\textwidth}
	   \centering
	   \includegraphics[width=1\textwidth,height=30mm]{{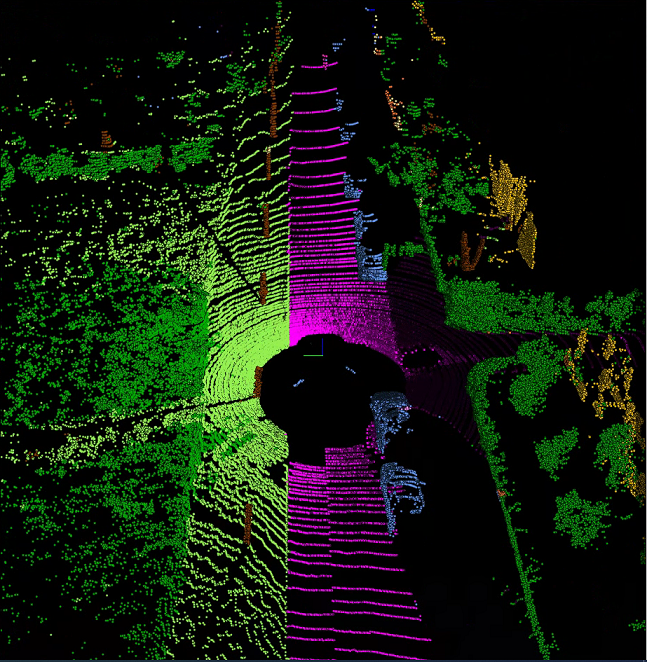}}
	\end{minipage}}
 \newline
  \subfloat[ground truth of scan 0]{
	\begin{minipage}[c]{0.32\textwidth}
	   \centering
	   \includegraphics[width=1\textwidth,height=30mm]{{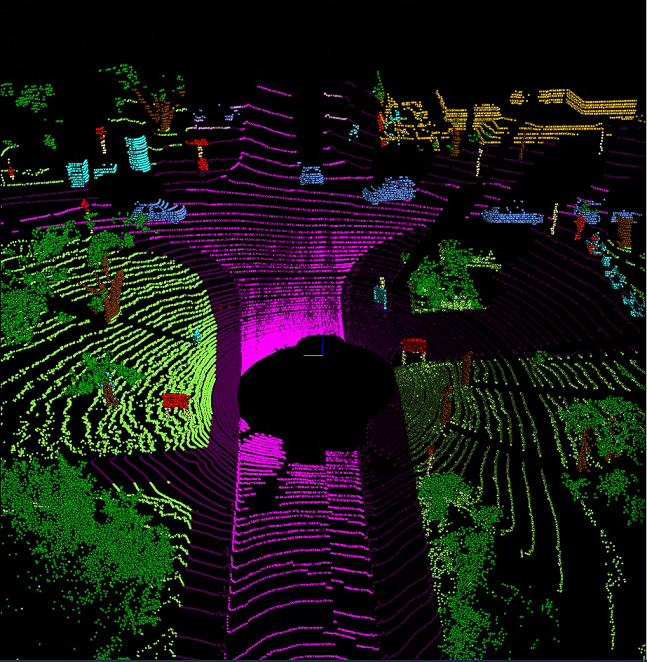}}
	\end{minipage}}
 \hfill 	
  \subfloat[ground truth of scan 40]{
	\begin{minipage}[c]{0.32\textwidth}
	   \centering
	   \includegraphics[width=1\textwidth,height=30mm]{{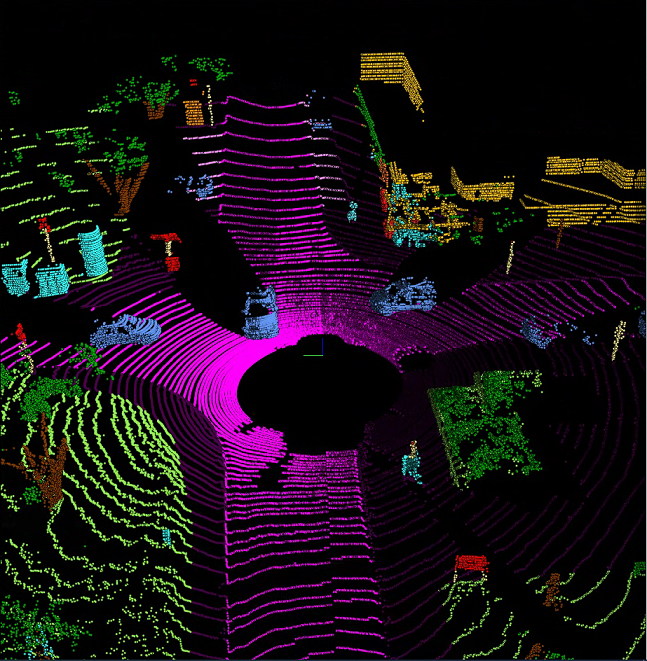}}
	\end{minipage}}
 \hfill 	
  \subfloat[ground truth of scan 140]{
	\begin{minipage}[c]{0.32\textwidth}
	   \centering
	   \includegraphics[width=1\textwidth,height=30mm]{{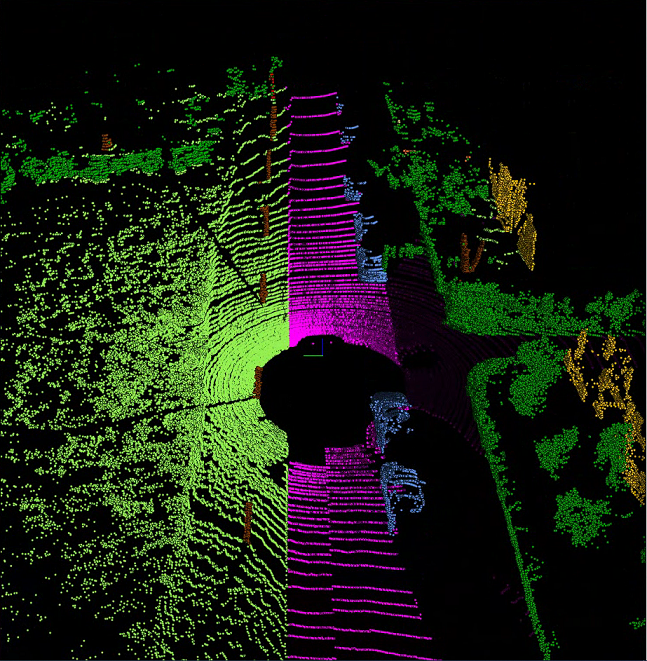}}
	\end{minipage}}
\caption{Comparison between our prediction and ground truth on validation dataset (sequence 08)}
\label{fig:result}
\end{figure*}

The performance of our network is illustrated in Tab.~\ref{tab:table1} and Fig.~\ref{fig:result}. Comparing to the SOTA network SalsaNext \cite{cortinhal2020salsanext} and FIDNet \cite{zhao2021fidnet}, our network shrinks the number of parameter to 1/4, about 1.4M. Besides, all the operations used in our network are DPU supported, which is not possible for other networks in the table.

\begin{table*}[htbp]
\begin{center}
    \caption{The performance comparison on SemanticKITTI \textbf{valid} set. All the listed networks are projection-based methods. (PolarNet result is from Table 3 of \cite{zhang2020polarnet}. SalsaNext is inferenced without uncertainty.)}
        \label{tab:table1}
    \setlength{\tabcolsep}{2.pt}
      \renewcommand{\arraystretch}{1.} 
 \begin{tabular}{c| c| c| c c c c c c c c c c c c c c c c c c c} 
 Methods & Size & \rotatebox{90}{\textbf{mean-IoU}} & \rotatebox{90}{car}& \rotatebox{90}{bicycle}& \rotatebox{90}{motorcycle}& \rotatebox{90}{truck}& \rotatebox{90}{other-vehicle}& \rotatebox{90}{person}& \rotatebox{90}{bicyclist}& \rotatebox{90}{motorcyclist}& \rotatebox{90}{road}& \rotatebox{90}{parking}& \rotatebox{90}{sidewalk}& \rotatebox{90}{other-ground}& \rotatebox{90}{building}& \rotatebox{90}{fence}& \rotatebox{90}{vegetation}& \rotatebox{90}{trunk}& \rotatebox{90}{terrain}& \rotatebox{90}{pole}& \rotatebox{90}{traffic-sign} \\ 
 \hline\hline
 
%
%
%
%
%
%
%
%
 
  SqueezeSegV3-21 \cite{xu2020squeezesegv3} & $64\times 2048$ & 51.0 & 87.0 & 31.4 & 48.9 & 24.7 & 33.6 & 49.8 & 59.0 & 0.0 & 93.0 & 37.0 & 80.0 & 3.0 & 85.1 & 40.3 & 85.0 & 52.1 & \textbf{73.1} & 47.1 & 38.2 \\
  
  SqueezeSegV3-53 \cite{xu2020squeezesegv3} & $64\times 2048$ & 52.7 & 86.1 & 30.9 & 47.8 & 50.7 & 42.4 & 52.2 & 52.4 & 0.0 & \textbf{94.5} & \textbf{47.3} & \textbf{81.6} & 0.3 & 80.2 & 47.2 & 82.5 & 52.5 & 72.0 & 42.4 & 38.2 \\
  
  PolarNet(Resnet-DL) \cite{zhang2020polarnet} & [480, 360, 32] & 53.6 & 91.5 & 30.7 & 38.8 & 46.4 & 24.0 & 54.1 & 62.2 & 0.0 & 92.4 & 47.1 & 78.0 & 1.8 & \textbf{89.1} & 45.5 & 85.4 & 59.6 & 72.3 & \textbf{58.1} & 42.2 \\
  
  SalsaNext \cite{cortinhal2020salsanext} & $64\times 2048$ & 55.8 &  86.2 & 39.4 & 42.0 & \textbf{77.7} & 42.0 & 62.1 & 68.3 & 0.0 & 94.3 & 42.2 & 80.0 & 4.1 & 80.0 & \textbf{48.4} & 80.3 & 57.9 & 64.2 & 46.6 & \textbf{44.5} \\

  FIDNet \cite{zhao2021fidnet} & $64\times 2048$ & \textbf{58.8} & \textbf{92.7} & \textbf{41.1} & \textbf{50.3} & 76.9 & \textbf{47.7} & \textbf{66.4} & 68.6 & 0.0 & 93.7 & 42.9 & 80.3 & 1.6 & 86.0 & 45.8 & \textbf{85.6} & \textbf{64.0} & 72.1 & 57.6 & 43.7 \\
 
  \textbf{Ours} & $64\times 2048$ & 56.4& 92.2& 37.5& 42.5& 72.4& 37.5& 63.2& \textbf{75.4} & 0.0& 92.0& 34.2& 77.7& \textbf{8.1}& 85.0& 45.3& 84.3& 58.7& 72.0& 50.3& 44.1\\
 \hline
\end{tabular}
\end{center}
    \vspace{-8mm}
\end{table*}

\section{DPU Implementation}
This network is further quantized and compiled targeting to Xilinx ZCU102 development kit using Vitis-AI workflow. The floating point model is first checked to ensure all the operators are supported by the DPU core. Then it is quantized into 8-bit representation. In this step, a subset of dataset is required for quantized weights refinement. In the last step, the quantized model is compiled to ZCU102 board. There are two DPUs implemented on the board, and the resource consumption is listed in Tab.~\ref{tab:resource}

\begin{table}[htbp]
    \begin{center}
    \caption{Hardware resources usage of a 2-core DPU}
    \label{tab:resource}
        \begin{tabular}{ |c|c|c|c| } 
        \hline
        FF & LUT & DSP & BRAM\\
        \hline
        203363 & 111565 & 1394 & 518 \\
        (37.1\%) & (40.7\%) & (55.3\%) & (56.8\%) \\
        \hline
        \end{tabular}
    \end{center}
\end{table}

The network pruning is discarded to compress our network. The main reason is, according to \cite{frankle2018lottery} and \cite{liu2018rethinking}, the network pruning has limited affect on light weighted networks. There is no difference on performance if pruning a trained heavy network or training the pruned network from scratch.

\begin{figure}[htbp]
    \centering
    \includegraphics[width=0.9\columnwidth]{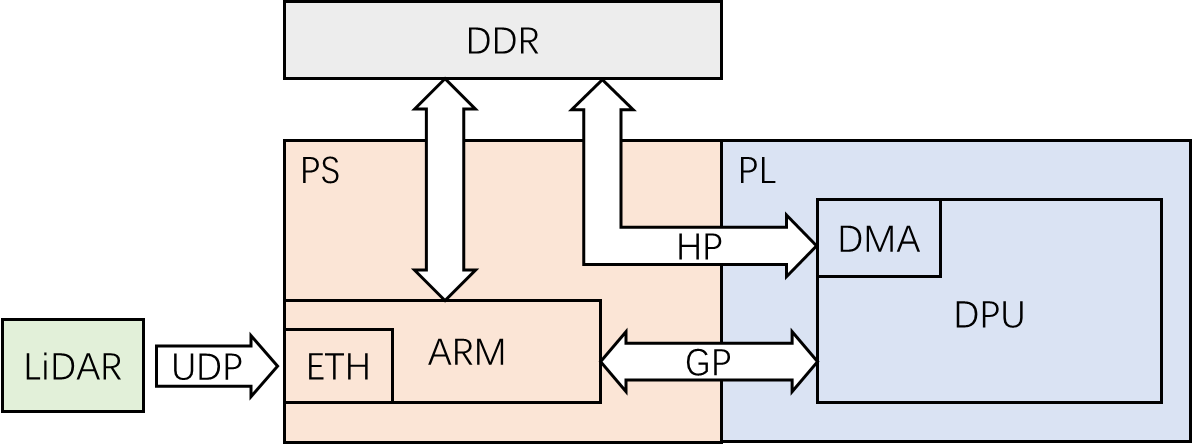}
    \caption{The system architecture of point cloud semantic segmentation system}
    \label{fig:system_arch}
\end{figure}

The system architecture is demonstrated in Fig.~\ref{fig:system_arch}. The LiDAR driver has been implanted into ZCU102 board and the ARM processor collects each set of point cloud into DDR for the DPUs on FPGA side. When running at 300MHz and splitting the 360\degree to 2 DPU cores, this system can process point cloud in real time (10fps). The power consumed is 16.8W. Therefore, the computation efficiency of our system is 42.5 GOP/W.

\section{Conclusion}
In this paper, a lighted weighted point cloud semantic segmentation network has been proposed. Different from existing networks running on GPUs, the limitation and computation efficiency of the edge deep learning accelerator has been considered during network design. All the operations in this network are fully supported by Xilinx DPU for edge applications. When tested on semantic KITTI dataset, it achieves 42.5 GOP/W in mIOU. If running on a 2-core DPU, it processes a 64-line dense point cloud at 10 fps.


%
\IEEEpeerreviewmaketitle

\newpage
\bibliographystyle{ieeetr}
\bibliography{semantic_seg.bib}

\end{document}